# Semantic Representation for Agentic Self-Play in Automated Laboratories

Yves Delley[1,†], John Coller[1], Zachary Cutts[1], Alexander Daryin[2], Mathias Bellaiche[1], Jun W Kim[1], Oscar Donosa[1], Srinidhi B Seshadri[2], Tanner Clary[2], Sandy Klemm[1], Avinatan Hassidim[1], Yossi Matias[1], Pooja Rao[1], Petar Sirkovic[3,†,‡] and Annalisa Pawlosky[1,†,‡]

[1]Google Research, [2]Google Cloud, [3]Google DeepMind, [†]Corresponding author, [‡]Equal leadership (co-senior authors)

**The complexity and dimensionality of biological search spaces represent a fundamental barrier to discovery. Many complex biological research questions require intelligent search strategies that examine numerous potential breakthrough hypotheses beyond known local minima in a scalable way. A direct approach designs combinatorial experiments across many hypotheses, and disentangles subtle biological signals from the noise. Biological experimentation in the areas of the unknown demands a scalable physical and computational intelligence to explore in tandem beyond local optima through iterative self-play across ideation, experimentation, and automation. The system would design iterative, data-driven experiments, and disentangle critical biological signals from multiple layers of noise across diverse experimental contexts. Translating experimental designs into reliable physical execution remains bottlenecked by static laboratory automation and unmodeled physical variability. To address this, we introduce VERified Infrastructure for Discovery Laboratories, or VERID, a foundational infrastructure designed to bridge the gap between the high-level reasoning of artificial intelligence (AI) agents and complex, error-prone physical realities of wet labs. VERID is a persistent, high-fidelity semantic digital twin of the laboratory workcell that provides four core services: queryable introspection for agentic reasoning, time-compressed simulation for protocol validation, comprehensive trace persistence for a complete scientific record, and a framework for safe execution.**

**We present three case studies using aqueous dye and deionized water mixtures to validate the optical and fluidic foundations of VERID. Together, these studies demonstrate how VERID provides the necessary interface to the wet lab hardware for robust agentic operation, enabling pre-emptive detection of critical errors in experimental protocols, stateful mid-run error recovery and post-hoc identification of subtle physical failures. The simulation functions as a discrete accounting engine, allowing agents to safely and inexpensively explore experimental designs *in silico* and refine experimental strategies through iterative self-play. By managing the complexity of wet lab experimentation setup and execution, VERID allows for robust structured multimodal experimental data collection, which is suitable for downstream analysis, knowledge mining and agent training. We argue that this semantic layer is an essential prerequisite for deploying autonomous agents in the biochemical and biomedical sciences, ensuring data integrity, operational safety, and paving the way for tackling previously intractable grand challenges in biology.**

## 1. Introduction

The process of scientific discovery from therapeutic drugs to novel materials and biological forms, has long been defined by high costs, long timelines, and immense complexity. The search requires solving two distinct, but intertwined problems: 1. navigating an immense, rugged search space and 2. physically executing laboratory experiments with high fidelity, adaptability and reproducibility. Progress in autonomous science has been limited so far because the two problems have largely been tackled independently, and pioneering closed-loop protein engineering platforms that couple them still rely on fixed, pre-programmed experimental workflows rather than adaptable agentic protocols [1].

The first challenge is the scale and structure of the biological search space. The number of possible drug-like small molecules is estimated to be as high as $10^{60}$ [2], [3], while a protein of just 100 amino acids has $20^{100}$ possible sequences, before even the consideration of post-translational modifications or non-natural amino acids [4], [5]. These spaces are too large to explore with linearly scaling methods or screening random libraries of $10^{14}$ molecules, such as phage,

*Corresponding author(s): delley@google.com, psir@google.com, apawlosky@google.com*

yeast or mRNA display [5], [6]. Additionally, starting positions from random libraries limit the exploration strategies early only, often resulting in local hill climbing results, rather than global explorations [5], [7]. The combinatorial explosion of possibilities renders traditional, human-led experimental methods statistically incapable of navigating these rugged, high-dimensional landscapes.

Recent advances in *in silico* modeling and LLM-driven agentic ideation offer a way to propose more informed starting points and hypotheses across these spaces [8]. In protein science, agents can prioritize candidate sequences by combining literature search with structural prediction and generative protein design tools, such as AlphaFold 3 [9], RFdiffusion [10], AlphaProteo [11], and BindCraft [12]. However, purely *in silico* exploration faces limits. Large language models remain prone to hallucination, and predictive models are not yet accurate enough, partly because they are trained predominantly on positive published outcomes, lacking access to negative data from failed experiments [13]. In-silico predictions also often fail to transfer directly to experimental reality. For example, even in state-of-the-art *de novo* protein binder campaigns strictly filtered by AlphaFold 2 [14] or AlphaFold 3 [9] metrics, 60% to $>$ 80% of computationally top-ranked designs typically drop off when tested experimentally via yeast surface display or downstream soluble biophysical characterization (e.g., failing to express, misfolding, or showing avidity-driven false positives that fail in biolayer interferometry) [10], [11], [12], [15], [16]. Without tight grounding in physical experimental feedback, purely computational ideation incurs significant resource loss.

The second challenge lies in how physical wet-lab experiments are run. Current laboratory practice is split between two paradigms: manual protocols and static automation. At the bench, researchers can customize manual protocols, but throughput is low, the work is labor-intensive, and prone to operator-to-operator variability [17]. Conventional laboratory automation achieves high throughput on static, pre-validated workflows. Translating a bench protocol onto a liquid-handling workcell requires specialized programming in proprietary, vendor-locked software, calibration of deck geometry and labware definitions, and empirical tuning of liquid-handling classes [18]. Thus, the breadth of protocols and experiments that can be scaled to robotic throughput and reproducibility is limited by the available time of domain experts, and stays limited to rigid protocols that do not adapt to new knowledge.

Recent work has sought to connect LLM-based reasoning agents directly to laboratory hardware to automate the design-build-test-learn loop in self-driving laboratories [19], [20], [21], [22]. Recent systems such as CLAIRify [23] and ORGANA [24] pair LLMs with compiler verifiers or task-and-motion planning to validate symbolic task plans, while open-source interfaces such as PyLabRobot [25] provide programmatic hardware control and in-memory volume and tip tracking within a single script. However, a critical operational gap remains for iterative agentic experimentation: existing frameworks either treat compiled unit operations as ideal or bind state to a single ephemeral script execution. They do not maintain a persistent digital twin that survives across modular command blocks and mid-run hardware faults, fork that live state for pre-dispatch simulation gating, or synchronize state transitions with video and assay readouts so agents can autonomously diagnose unmodeled fluidic and optical biases. Without a persistent intermediary layer that gates commands against live workcell context, an agent can generate syntactically valid code that is physically impossible (e.g. lowering an 8-channel manifold into a narrow trough), logistically invalid (e.g. aspirating from a depleted well or transferring into a lidded plate), or scientifically compromised by unmodeled fluidic effects. Whether a command is safe and effective cannot be determined from the code alone, but depends on the live physical context of the workcell, such as current meniscus heights or prior failure states. An intermediary layer that tracks and validates physical state across runs is critical so that autonomous agents do not operate on flawed assumptions, produce confounded data, or physically crash instruments. Furthermore, accurate state tracking is essential for self-improvement, auditing and scientific reproducibility.

To bridge computational ideation and reliable physical execution across this three-layer hierarchy, we introduce **VER**ified **I**nfrastructure for **D**iscovery Laboratories, or **VERID**, a semantic digital twin architecture designed to ground high-level reasoning agents in the physical laboratory. Unlike a static simulator which only predicts what could happen in a proposed protocol, a digital twin maintains rich context via a persistent, bidirectional link to the physical system it represents, tracking what has happened across experimental runs and using it to improve future predictions [26]. Here, we describe the architecture and four core services of VERID: (a) queryable introspection, (b) time-compressed simulation, (c) multimodal trace persistence, and (d) safe execution, and demonstrate through foundational fluidic and optical physics benchmarking wet-lab case studies how this semantic layer enables pre-execution error prevention, stateful error recovery, and causal debugging of subtle physical artifacts. The case studies we present were carefully designed to cover the most fundamental laboratory operations ubiquitous in biological bench research including aqueous reagent tracking, transfer via pipette, mixing and incubating. These benchmarks allow high-efficiency development of agentic wet lab capabilities by avoiding expensive biochemical reagents and instead leveraging low-cost experiments using dyes [27] to enable fast quantitative system characterisation, before composing lab operations into more sophisticated experiments and deploying on open biological problems.

## 2. Results: The Semantic Agentic Lab Twin (VERID)

VERID acts as the foundational interface between high-level reasoning agents executing the experiments and the physical robotic workcell. It abstracts low-level hardware communication while exposing a queryable, controllable model of the wet lab. By design, VERID separately abstracts the concepts of physical lab equipment and experimental protocols, while always persisting general state and collating multimodal data feeds, enabling automation to scale easily across different labs, robotic hardware and consumables unlike automations written on proprietary platforms. The core of VERID is its persistent digital twin of the workcell, which extends the in-memory resource hierarchy and volumetric/tip trackers of PyLabRobot (PLR) [25] into a long-lived, server-managed state engine across independent command blocks and instruments. This digital twin maintains a live, queryable entity tree capturing the spatial coordinates, container geometry, operational status, and semantic contents (e.g. reagent identity, volume, and liquid height) of every item on the deck, from robotic arms and plate readers to individual wells and pipette tips. Using this ontological hierarchy, which formally defines containment and spatial relationships, such as a well belonging to a plate or a lid obstructing well access, VERID manages the execution of command blocks — modular programs representing discrete stages of an experimental protocol. A command block can be a shorter protocol such as a dilution series, or subsections of longer protocols. Every command block must pass a simulation gate before physical dispatch, allowing agents to verify feasibility and analyze in-depth different protocol designs *in silico* before committing physical resources, allowing for deeper customization and exploration while simultaneously saving resources.

### 2.1. Adaptation for Agentic Reasoning

VERID exposes the digital twin through a structured, queryable API that allows an agent to inspect the live workcell state. For example, VERID can ask questions such as *"What is the current volume and liquid height in well C3?"* or *"Which pipette tips and reagent troughs have sufficient capacity?"*. This inspection provides the situational awareness needed for an agent to generate context-aware instructions grounded in physical constraints, rather than attempting to infer deck state from unstructured text logs or raw camera pixels.

### 2.2. Efficient Simulation for Exploration

Leveraging the semantic digital twin, VERID provides a time-compressed, *in silico* simulation engine that functions as a deterministic discrete-state accounting system. Before any physical action is deployed, the controller clones the current state of the digital twin and executes the agent's proposed command block against the virtual replica in a mandatory virtual run.

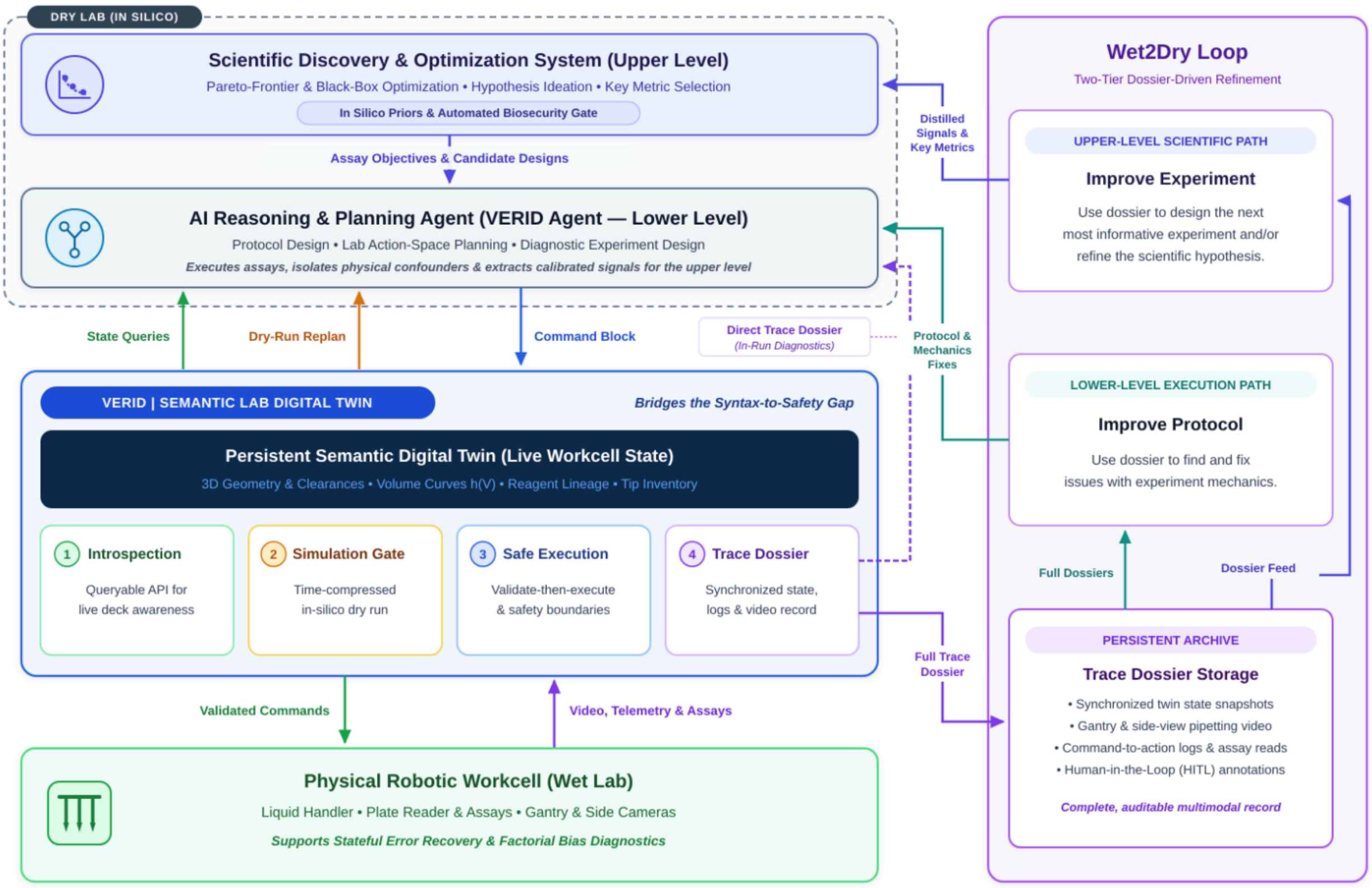

Figure 1 | Architecture of VERID and closed-loop integration between computational ideation (Dry Lab) and physical execution (Wet Lab). Within the dry lab, an upper-level Scientific Discovery & Optimization System proposes hypotheses and candidate designs, which a lower-level AI Reasoning & Planning Agent translates into modular experimental protocols. VERID bridges the syntax-to-safety gap via a persistent semantic digital twin of the live workcell and four core services: (1) Introspection, a queryable API for live deck awareness; (2) Simulation Gate, time-compressed *in silico* dry runs; (3) Safe Execution, validate-then-execute hardware dispatch; and (4) Trace Dossier, synchronized state snapshots, execution logs, and multi-view video. Full trace dossiers are archived into the Wet2Dry Loop, which uses the stored dossier along two feedback paths: an upper-level path to improve the experiment (designing the next most informative experiment and/or refining the scientific hypothesis) and a lower-level path to improve the protocol (finding and fixing issues with experiment mechanics across runs).

The simulator tracks mass balance, tip state, deck geometry, and elapsed time, fast-forwarding through incubations and waiting periods. This lightweight representation catches a wide spectrum of failure modes prior to execution, such as logistical errors (e.g. pipetting into a lidded plate), resource exhaustion (e.g. depleting pipette tips or reagent stocks), geometric collisions (e.g. channel pitch exceeding container bounds), and invalid volumetric parameters (e.g. aspirating air from an empty well or overflowing a destination well). Agents can therefore use the simulator as a high speed testing sandbox and refine candidate protocols without consuming reagents.

## 2.3. Trace Persistence for a Complete Scientific Record

Across every protocol run, VERID compiles a comprehensive research dossier by synchronizing three multimodal data streams against a common reference clock. First, it captures full digital twin state snapshots before, during, and after every command block, preserving modeled well volumes, tip inventories, and labware positions across runs. Second, it records time-tagged top-down gantry imagery, side-view pipetting video, and instrument telemetry aligned to the execution timeline. Third, it maintains a line-by-line execution log linking high-level protocol code directly to low-level hardware commands, state transitions, and video frames, providing a complete, auditable record for downstream debugging, reproducibility, and agent learning.

### 2.4. Safe Execution

While simulating a command block, VERID evaluates mechanical trajectories, volumetric bounds, and chemical compatibility rules against the current digital twin state prior to dispatch, and filters out unsafe or destructive actions before they reach the physical hardware, directly addressing the syntax-to-safety gap. This simulation is continuously re-synced as information flows from the physical hardware to the digital twin, and re-gated in a verify-then-execute safety framework.

## 3. Results: Case Studies

In order to test the VERID intelligence harness, we conceived of three case studies that illustrate proof-of-principle automated protocol debugging, multi-step fault recovery, and closed-loop experimental self-play – all prerequisite capabilities for robust agentic operation. To avoid biological complexity and focus on tool performance, all case studies utilise dye and deionized water mixtures [27] – allowing for fast experiment cycles, low reagent costs, and clear performance metrics, while demonstrating experimental conditions comparable to many essential steps in wet-lab liquid handling. The case studies present different contexts in which an agent interacts with VERID as follows:

1. **Human in the Loop protocol refinement:** The agent was tasked with a characterisation task via an interactive agentic interface. When the agent detected a discrepancy between expectation and reality, it reported its findings together with a hypothesis. The human operator then identified the root cause, enabling the protocol to be fixed.
2. **Supervised multi-step diagnosis:** An agent was tasked to develop a repeatable wet-lab protocol. The agent autonomously identified several traps and issues in a naive protocol implementation. The human operator only guided which limitations to accept and which ones to attempt to overcome. For the latter, the agent autonomously performed root-cause analysis.
3. **Unsupervised self-learning via challenge solving:** The agent was asked to solve a specific task using fixed, limited resources, and to allocate these resources wisely. It was also asked to write a record of transferrable learnings obtained during the execution. The experiment was repeated with varying boundary conditions, such as having access to previous learnings or not, or being allowed to look at intermediate results before dispatching further lab tasks.

### 3.1. Case Study 1: Human-in-the-Loop (HITL) Refinement of Serial Dilution Fluidics and Stateful Error Recovery

To illustrate how VERID supports iterative protocol debugging, we look at a case where an agent was tasked with characterizing the absorbance linearity of four dyes (red, yellow, blue, and green, each aspirated in duplicate from a reagent trough and a Falcon tube) across an 11-step, 2-fold serial dilution in a 96-well plate. In the initial execution, the agent relied on default well-bottom pipetting coordinates ($z = 0$ mm) and five in-place 100 µL mixing cycles at a fixed height. After reviewing the absorbance measurement data obtained from the plate reader, the agent had observed both severe systemic under-dilution and large inter-replicate spread: instead of the target 2.00× concentration reduction per column ($\log_2$ slope $= -1.00$), the measured step dilution factor was only ∼1.4×–1.5× for red, green and yellow, and 1.6×–1.8× for blue. The agent had hypothesized insufficient mixing to be the cause. The operator then reviewed synchronized video showing the side of the transparent wells of the assay plate, cross-referencing the execution trace and physical well geometry, which exposed two compounding physical failure modes at $z = 0$ mm:

1. dispensing, mixing, and re-aspirating at the well floor left dense dye stratified at the bottom so that ∼68–70% of the dye was carried forward at each transfer step instead of 50%
2. minor channel-to-channel height tolerances caused individual tips to press flush against the flat well floor and partially occlude the tip orifice during dispensing and mixing, driving the large discrepancy between replicates – visible to the human eye as a complete lack of dye being introduced until the tip was lifted off from the well bottom again.

In response, the agent refined the protocol to enforce a 1.5 mm bottom clearance on all liquid transfers and perform vertical convective mixing across the filling range (aspirating 100 µL near the bottom at $z = 1.5$ mm and dispensing near the top surface at $z = 4.8$ mm over four cycles per column). When this refined run encountered an unexpected instrument-state error after completing all liquid-handling steps, VERID's persistent digital twin state allowed the agent to resume in a follow-up cycle without re-pipetting: the agent moved the prepared plate out of the reader, executed an empty-chamber calibration, and completed the multi-wavelength readout.

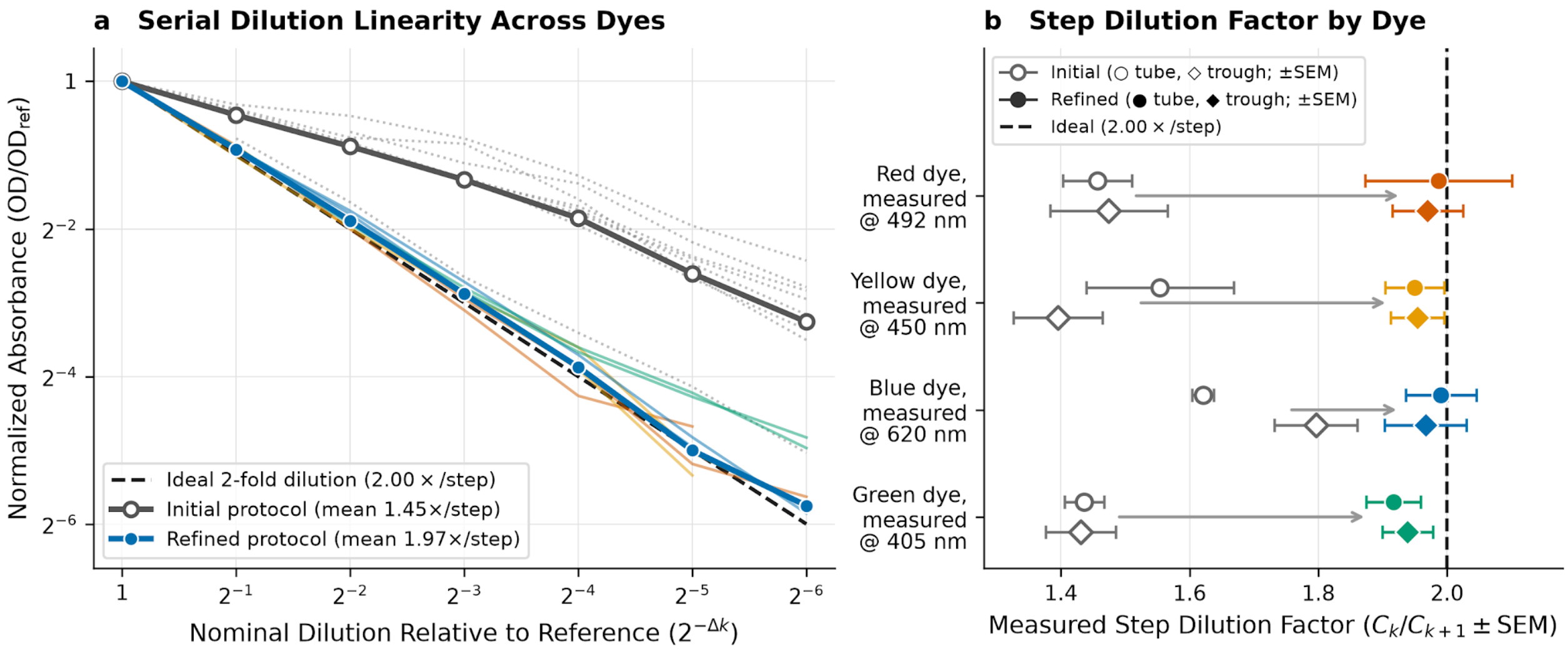

Figure 2 | Panel **a** overlays the mean normalized unsaturated dilution trajectories (legend) over all 8 dye–source series (dotted: initial protocol, thin lines colored by dye: refined protocol), and Panel **b** shows the per-dye, per-source step dilution factor ($\frac{C_k}{C_{k+1}}$, target 2.00×).

Across the unsaturated linear regime, the refined protocol eliminated the inter-replicate spread ($\leq 0.02\times$ difference between tube and trough across all four dyes) and restored near-ideal 2-fold dilution, achieving average step dilution factors between 1.92×–1.99× (see Figure 2). This case study demonstrates how VERID's persistent digital twin and multimodal trace persistence provide the granular operational data necessary for operators to identify and isolate subtle mechanical sources of systematic experimental error.

### 3.2. Case Study 2: Simulation-Gated Protocol Planning and Multi-Step Diagnosis

In our second case study, the agent was tasked with standardizing four concentrated dye stocks in a 12-column reagent trough each to a specific (per-dye) target optical density at their respective strongest absorbance wavelength as visible to our plate reader. Whereas Case Study 1 examined single-operation calibration, this second case illustrates how the persistent semantic model (combining container geometry, volume-to-height curves $h(V)$, tracked inventories, and pre-execution simulation), guides multi-step planning, mid-run recovery, and root-cause diagnosis (Figure 3).

Before execution, the model correctly identified that the trough capacities (15.7 mL) combined with the target reagent volumes (12 mL) would make in-place dilution impossible without determining stock concentration separately before mixing, given an uncertainty spread of stock dye exceeding the headroom ratio. Because accurate concentration measurement is comparatively involved (the agent selected a three-level dilution series), the agent instead approached standardisation with a plan of top-up, measure, remove, dilute. The controller system also caught an unphysical proposed operation – when the agent attempted to lower all channels of the Span-8 into a trough simultaneously, the simulator rejected the move as 8 tips span at least 72.0 mm, exceeding the 71.2 mm trough length (Panel a). Within the same turn, the agent adjusted its plan to mix the trough using 7 channels (Channels 1–7, 63.0 mm span) and sample in two sequential 4-channel passes (Channels 1–4, then 5–8).

During execution of the serial dilution, a mid-run communication failure interrupted the protocol on the fourth dye (Panel b). The ability of the digital twin to persist state allowed the system to continue the run from the fault point instead of re-running all dyes anew, as it had preserved the virtual state corresponding to the completed wash, blank read, and serial dilution of the first three dyes.

Post-run evaluation, however, revealed a 15–20% concentration discrepancy between adjacent dilution levels of the serial dilution and a 28–47% absorbance drop on Channel 8 across all dyes. To diagnose this experimental failure, the agent halted the protocol and autonomously designed an 8-column factorial diagnostic split across two runs, filling and blanking all 12 columns in Run 1 so Run 2 resumed directly on persisted Columns 5–8 without setup overhead. This diagnostic (Panel c) showed flat delivery across all 8 channels when using dry tips (1.58 OD, CV = 1.4%), ruling out geometric or meniscus-tracking errors. Replaying the 7-channel trough mix (Channels 1–7 mixed at 250 µL, Channel 8 dry) reproduced the apparent −27.7% drop on Channel 8 (1.63 OD), which matched the true dry-tip baseline. Crucially, it also ran a reverse 7-channel mixing assignment to Channels 2–8 (leaving Channel 1 dry), which restored Channel 8 (2.20 OD) and shifted the −28.6% deficit completely to Channel 1 (1.58 OD). Rather than Channel 8 under-aspirating, the 250 µL trough mix coated the inner wall of the 7 mixing tips with concentrated dye that washed down during the 150 µL well mix, adding +4.9 µL (+40.6%) of carryover, which also quantitatively explained the 14.5% offset between the 24 µL and 12 µL dilution levels.

The digital twin capabilities enabled the agent to discover the root cause of experimental error systematically and in the same way a human laboratory scientist would approach the issue – namely by observing the key measurable quantities, hypothesizing about the root cause (scientific or ‘hardware’-based), and iteratively adjusting the experiment, until a clean measurement is reached.

### 3.3. Case Study 3: End-to-End Physical Self-Play in Automated Laboratories

In the third case study, we evaluated whether an autonomous agent can characterize an unfamiliar physical assay, diagnose unmodeled instrument artifacts, and improve its experimental accuracy across within-run feedback cycles and multi-generation runs using only its own physical observations (Figure 4). In each challenge (Figure 4 a), an automated Challenger prepares four blinded binary dye mixtures at unknown concentrations alongside four pure reference stocks (Red, Green, Blue, and Yellow). Operating under a fixed physical budget (one 96-well plate, a capped pipette-tip allowance, and overhead RGB camera imaging only), a blind Solver agent designs and executes closed-loop experimental cycles – serial dilutions, synthetic reference mixtures, or standard-addition spikes – to infer each mixture's composition. After the run, an independent four-wavelength plate reader (Byonoy Absorbance 96; Oracle) measures sealed reference aliquots to measure the target composition and the exact source dye concentrations exactly, and then scores the Solver's probabilistic predictions against the true mixture via a tempered KL-divergence expressed in millinats (mnats; lower is better, zero is exact recovery). The reference measurement allows us to separate small variations in the target mixture preparation from errors the agent made in determining the composition. We compared Single-Cycle (SC) and Multi-Cycle (MC) solvers across three paired challenges: Challenge 1, Challenge 2A, and Challenge 2B (Figure 4 b). Solvers never access ground-truth identities or Oracle scores; between Challenge 2A and Challenge 2B, only a sample-blind summary of general physical and procedural learnings is passed to the next generation.

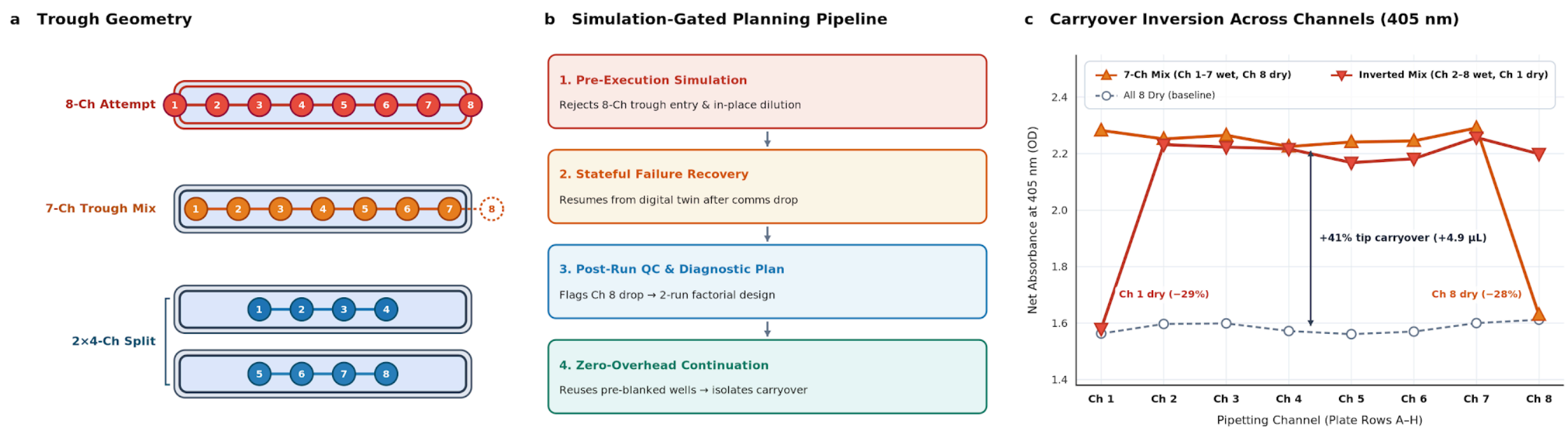

Figure 3 | (a) Trough geometry constraint: simultaneous 8-channel entry collides with the reagent trough walls, prompting the agent to mix with 7 channels (leaving Channel 8 outside and dry) and sample in two sequential 4-channel passes (Channels 1–4, then 5–8). (b) Four-stage planning and execution pipeline across pre-execution simulation, stateful recovery after a mid-run communication failure, post-run QC and factorial diagnostic design, and zero-overhead continuation on pre-blanked wells persisted in the digital twin. (c) Net absorbance at 405 nm (12 µL Green dye) across Channels 1–8. Dry-tip baseline is flat (1.58 OD, CV=1.4%). Inverting the 7-channel trough mix from Channels 1–7 (orange ▲, Ch 8 dry) to Channels 2–8 (red ▼, Ch 1 dry) shifts the −28.6% deficit from Channel 8 to Channel 1, isolating the +41% (+4.9 µL) wall-film carryover added during well mixing.

Depending on the pair of dyes selected for a particular target mixture, the difficulty of identification varies significantly. Beer–Lambert absorbance of monochromatic light is linear in the concentration. With only three channels available on the camera (RGB Bayer filters), uniquely inferring four concentrations in the linear regime is not possible. For the dyes used in these experiments, the apparent RGB absorption coefficient vectors of blue, green and yellow dyes approximately fall within a two-dimensional subspace, whereas the RGB absorption coefficient vector of the red dye is orthogonal to the rest. The challenge is well-posed because the Bayer filters are broad-band, and thus multiple wavelengths contribute. The exponential connection between absorption coefficient and transmitted light-intensity provides the necessary nonlinearity to enable separation in principle, but with stark variations in difficulty. A successful experiment design must exploit this nonlinearity by affecting the effective absorbance of different dyes non-uniformly, by "spiking" the unknown mixture with targeted additions, while navigating the many metrological challenges around signal saturation, noise floor, automatic white-balance, viewing-angle dependent reflection and refraction, as well as the experimental challenges of accurately transporting small volumes of liquids, dye segregation, evaporation, surface effects or non-Beer–Lambert sources of brightness changes such as backscattering or fluorescence.

We therefore expect that an agent will need to identify patterns of these effects that may only develop over many experiment cycles, as well as carefully adapt its limited resources to the specific target mixtures it is facing. We posit that these challenges are representative of at least some of the challenges encountered in biochemical assays.

Following the multi-cycle lineage from MC-2A to MC-2B (Figure 4 c) shows how cross-run learnings convert early experimental errors into effective multi-cycle exploration. Starting cold on Challenge 2A, MC-2A attempted to resolve ambiguous mixtures in later cycles by spiking unknown wells with small volumes of concentrated pure stocks. These spikes saturated the camera's red channel and introduced small-volume pipetting variance, degrading overall loss from 247 to 307 mnats across four cycles even as its orthogonal Red+Blue prediction improved. Without access to true mixture identities, MC-2A diagnosed both the camera saturation ceiling and small-volume pipetting jitter from its own curve-fit residuals and recorded strict operating bounds in its learnings. Inheriting those bounds on Challenge 2B, MC-2B redesigned its two-cycle strategy: a sparse initial dilution survey with pre- and post-dispense blank imaging in Cycle 1 (137 mnats), followed in Cycle 2 by four-point serial dilutions of its top candidate pairs at matched concentrations. Comparing full dilution trajectories rather than single spiked wells corrected a Cycle-1 miscall on Red+Green, refined Yellow+Green, and calibrated uncertainty across the Yellow+Blue twin pair, reaching 58 mnats mean loss, a more than five-fold improvement over MC-2A. By contrast, without prior learnings (Challenge 1), MC-1 (85 mnats) outperformed SC-1 (205 mnats) through within-run feedback alone, whereas the single-cycle lineage SC-2A→SC-2B, 142→162 mnats stalled because committing the entire experiment in one blind shot left SC-2B with single-concentra-tion validation wells that remained degenerate between twin pairs.

Across Challenge 2A and Challenge 2B, agents progressed from correcting basic geometric and fluidic artifacts to modeling optical physics purely from internal consistency residuals (Figure 4 d). First, agents detected that oblique camera viewing angles caused lit well walls to bleed into outer-row measurements, and used the 3D camera calibration to project well rims and floors separately, restricting measurements to parallax-free floor disks while fitting a smooth 2D illumination surface across the plate. Second, from jumps in replicate residuals, agents identified that aspirating nearly empty source wells introduced air bubbles, small stock transfers suffered high volume variance, and concentrated mixtures clipped the camera's 8-bit channels, prompting minimum transfer and residual-volume bounds alongside wet blanking. Third, agents replaced channel-by-channel Beer–Lambert fits with a linear-light model incorporating cross-channel camera coupling, optical flare, and the negative apparent red-channel absorbance of Red dye. Finally, in Challenge 2B, agents detected vertical concentration gradients in unmixed deep-well source plates and derived the linear-light twin degeneracy among Yellow, Blue, and Green, motivating the multi-point dilution trajectories and calibrated probability hedging used to resolve degenerate pairs.

## 4. Discussion

The use cases we described here serve as empirical validation of the core services of VERID, and showcase how our approach connects high-level scientific reasoning with automated wet-lab execution by establishing robust physical and informational infrastructure. Crucially, this architecture addresses the two primary barriers to efficiently navigating rugged biological search spaces: isolating and eliminating systematic experimental bias during physical assay execution, and closing the computational-experimental loop via structured trajectories and autonomous learning.

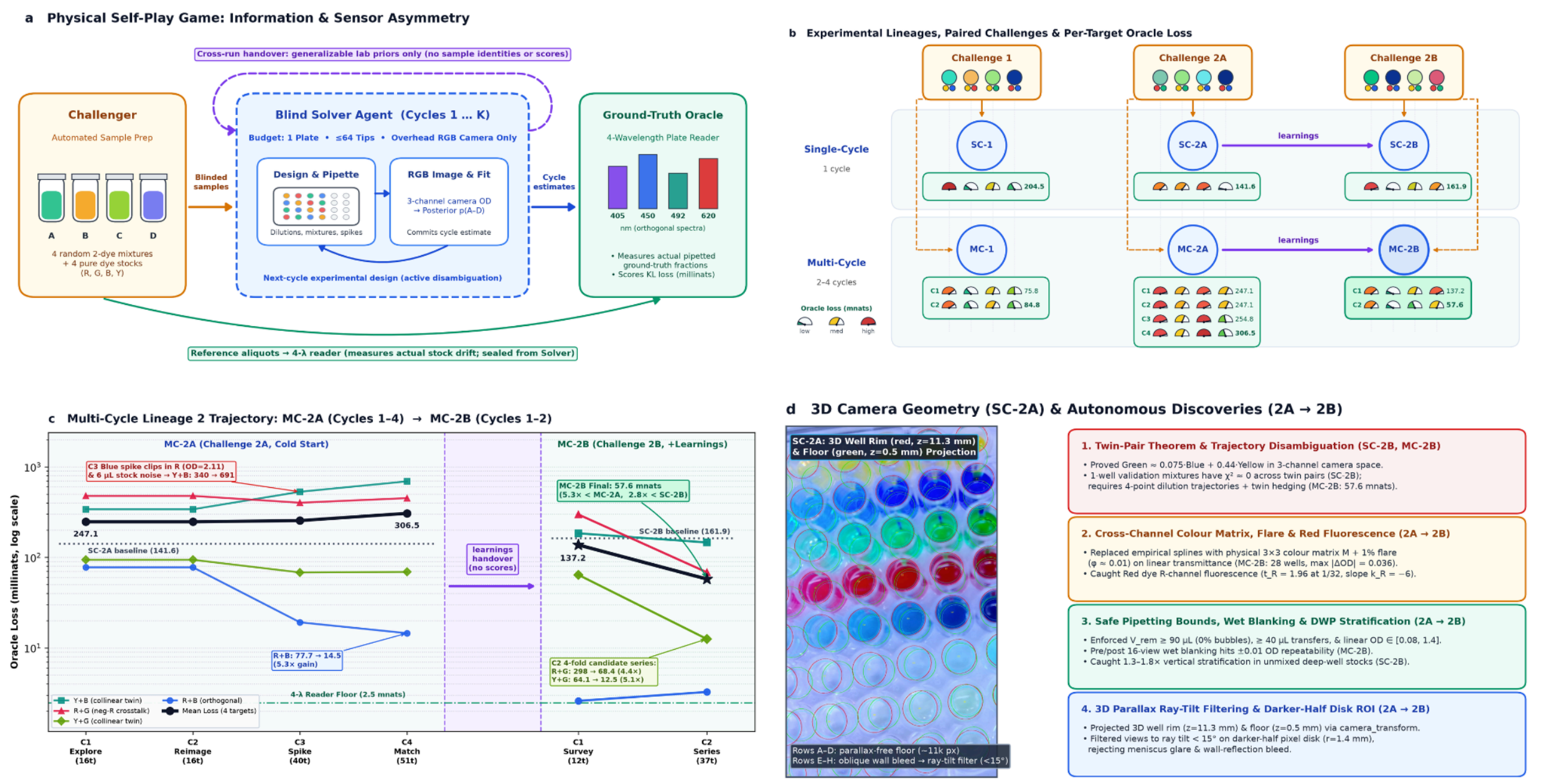

Figure 4 | End-to-end physical self-play in automated laboratories. (a) Physical self-play setup: an automated Challenger prepares four blinded binary dye mixtures alongside four pure reference stocks (Red, Green, Blue, and Yellow). Under a fixed physical budget (one 96-well plate, capped tip allowance, and overhead RGB camera imaging only), a blind Solver designs and executes closed-loop experimental cycles to infer each mixture's composition, while an independent four-wavelength plate reader (Oracle) scores probabilistic predictions on sealed reference aliquots in millinats (mnats). (b) Experimental lineages across Challenge 1, Challenge 2A, and Challenge 2B comparing Single-Cycle (SC) and Multi-Cycle (MC) solvers, with per-target Oracle loss across spectrally orthogonal (Red+Blue, Yellow+Red) and spectrally degenerate or crosstalk dye pairs. (c) Multi-cycle lineage trajectory from MC-2A (Cycles 1–4, cold start) to MC-2B (Cycles 1–2, with inherited sample-blind learnings), showing how operating bounds diagnosed from internal residuals in MC-2A (307 mnats final loss) enable a two-cycle dilution survey and trajectory comparison in MC-2B (58 mnats mean loss). (d) Autonomous physical and optical discoveries derived from internal consistency residuals across Challenges 2A and 2B, including 3D parallax-free well-floor projection, safe pipetting and dynamic-range bounds, cross-channel coupling with Red-channel brightening, and three-dye linear-light twin degeneracy.

### 4.1. Isolating Systematic Experimental Bias in High-Throughput Assays

In experimental biology, it is important to distinguish between unbiased stochastic noise, systematic experimental bias and true biological signal. Unbiased random error can be simply mitigated through statistical replication, while systematic bias is more pervasive in high-throughput discovery and confounded by instrument tolerances, labware geometry, fluid properties, and subtle, untracked variations in protocol execution [13], [17]. Large-scale combinatorial platforms, such as DNA-encoded libraries (DELs), yeast surface display, or massively parallel reporter assays (MPRAs), generate high-dimensional readouts where physical execution artifacts easily masquerade as biological signal or assay system preference, leading models astray when trained on uncalibrated outputs.

An agent operating through a semantic digital twin is uniquely positioned to detect, diagnose, and suppress these systematic confounders by explicit tracking of execution covariates and closed-loop factorial diagnosis. Conventional automation logs only whether a script completed, discarding the physical details of how each well was handled. By recording the full execution trace (including such varied measurements as channel assignments, tip reuse history, aspiration heights $z(t)$, meniscus clearances, dwell times, and synchronized video), VERID turns hidden execution mechanics into explicit, queryable covariates. When an assay exhibits spatial gradients or batch discrepancies, the agent can correlate anomalies directly against physical execution parameters rather than treating the variation as unexplained biological noise. When a systematic discrepancy is detected, the persistent digital twin allows the agent to pause a campaign and execute targeted factorial controls on the exact same deck state to isolate the physical root cause. Case Studies 1 and 2 illustrate this capability directly: in neither case could simple replication have fixed the error. In Case Study 1, replicate transfers from $z = 0$ mm consistently under-diluted due to density stratification and tip-floor occlusion; in Case Study 2, repeating the 7-channel trough mix reproducibly biased Channels 1–7 by $+40.6\%$ via inner-wall film carryover while making Channel 8 appear defective. By inspecting the trace and inverting channel assignments in a controlled diagnostic run, the agent isolated the fluidic mechanism and corrected the protocol before downstream experimental campaigns were compromised.

### 4.2. Data Synchronisation and Persistence for Knowledge Extraction

Once physical execution is calibrated and auditable, the multimodal datasets captured by the Trace Persistence service provide the foundation for closing the loop between physical experimentation and *in silico* exploration. VERID synchronizes all heterogeneous data streams, digital twin state snapshots, command execution traces, video feeds, hardware telemetry, and endpoint assay readouts onto a unified timeline and outputs it as a single trace dossier. This harmonised experiment-specific data layer can then be reasoned over agentically to enable iterative discovery science via e.g. improved design of experiments, refinement of originally generated hypotheses or even reinforcement learning of the underlying AI model. Automating large-scale indexing and knowledge extraction across archived trace dossiers—and systematically benchmarking its impact on downstream campaigns—remains future work.

### 4.3. Closing the Loop for Self-Play Agentic Learning

Case Study 3 demonstrates how agents can autonomously explore the physics of wet-lab robotics and accumulate transferable expertise: an agent identifies gaps in its assumptions and patterns of unexpected physical behavior, distilling key operational rules and observation models to initialize subsequent iterations. This extraction can be performed inline by the experiment-driving agent (as in Case Study 3) or asynchronously by a dedicated analysis agent auditing historical trace dossiers. Crucially, this learning loop does not require a task-specific extrinsic reward function. Instead, the learning signal is the self-supervised prediction error—the mismatch between expected and observed physical outcomes—which can be compared to the world-model concept extensively discussed in other domains of robotics and reinforcement learning [28], [29], [30], [31].

However, whereas conventional robotic world models typically train deep neural networks to predict high-dimensional future sensor readings (such as raw video pixels) directly from past frames and actions [28], [29], an agent coupled to VERID operates at a semantic and symbolic level [30], [32], [33]. Because the digital twin already enforces known volumetric conservation laws and geometric constraints, the agent only needs to learn the unmodeled physical residuals—such as tip-wall film carryover, density stratification, or optical cross-coupling. In our current implementation, these discoveries are persisted as structured natural-language operating bounds and analytical models in an experiential knowledge base [34].

Grounding autonomous agents in an explicit, refinable world model enables more principled trade-offs between exploration and exploitation than black-box trial-and-error optimization. Rather than deploying unguided parameter sweeps, an agent can inspect which physical assumptions its candidate protocol relies upon, exploit verified operating regimes directly, and execute targeted factorial diagnostics (as demonstrated in Case Study 2) to resolve contested parameters in-line.

### 4.4. Closing the Loop for Biological Hill-Climbing

Enabling agents to autonomously perform wet-lab quantification in a repeatable and accurate manner provides a direct way to hill-climb on biological problems—such as transitioning from singleton screens to synergistic combinations in drug repurposing, dosing intrinsically disordered proteins (IDPs) with small-molecule pools to induce structure, or guiding cellular perturbations through live-cell microscopy—by feeding the measured biological objective back into a domain-adapted or general black-box optimization procedure. However, with the full trace dossier available, a suitably sophisticated optimization algorithm can also leverage details of the experimental execution when selecting the next step to try. This can include intermediate optical or imaging readouts that correlate with the terminal objective or flag upstream sample-preparation issues (such as early compound precipitation/turbidity or transient sub-cellular morphological changes), as well as tracked procedural covariates that correlate with systematic bias and measurement uncertainty (such as incubation dwell times, unsealed evaporation exposure, labware-surface contact history driving non-specific adsorption, or spatial plate and channel assignments).

### 4.5. Safety and Responsible Innovation

Deploying autonomous agents in a physical laboratory requires a multi-layered safety architecture encompassing operational hardware safety, human oversight, and biosecurity.

#### *4.5.1. Operational Safety*

To close the syntax-to-safety gap, VERID's execution layer draws on formal safety frameworks for self-driving laboratories, such as Safe-SDL [35]. Verified Operating Design Domains (ODDs), high-level operational boundaries such as container volume capacities, geometric clearances, maximum agitation speeds, and chemical compatibility matrices, are encoded as machine-checkable invariants within the digital twin and enforced at the pre-execution simulation gate [35], [36]. Mechanical models prevent gantry trajectories that risk hardware collisions or cross-contamination, while chemical constraints reject reagent combinations that could produce hazardous reactions.

#### *4.5.2. Human-in-the-Loop (HITL) Review*

Human judgment remains essential for interpreting ambiguous experimental phenomena and overseeing high-level scientific direction. The synchronized trace dossiers enable human review by leveraging the digital twin to render synchronized, human-readable reconstructions of every run, aligning line-by-line code execution with gantry video and well-level state histories. Scientists can inspect anomalous steps, diagnose physical artifacts, and record structured annota-tions. These expert corrections feed directly back into the training archive, systematically capturing tacit laboratory know-how in machine-readable form.

#### 4.5.3. Automated Biosecurity

When agentic workflows involve *de novo* nucleic acid or protein synthesis, special regard must be taken to prevent accidental or deliberate generation of known or novel hazardous biochemicals and pathogens. Extending the mandatory verification gate with compositional lineage tracking and physical-chemistry propagation rules will further enable automated detection of *de novo* DNA and RNA synthesis protocols. An auto-reviewer can then flag these protocols for screening via an automated biosecurity gateway, integrating function-based sequence screening tools such as SecureDNA [37]. Because generative models can design functional sequences with low sequence identity to known pathogens, potentially evading traditional homology-based screens, cryptographic and function-aware screening provides an essential safeguard.

## 5. Conclusion

Overcoming grand challenges in biology, from engineering bespoke enzymes and high-affinity therapeutics to mapping complex regulatory networks in cellular systems, requires both intelligent navigation of vast combinatorial spaces and rigorous, reproducible physical execution. Neither computational ideation nor traditional laboratory automation is sufficient on its own. *In silico* predictions can suffer from high experimental drop-off without physical grounding.

VERID mends the divide by placing a persistent, queryable semantic digital twin between the reasoning agent and the physical workcell. As our experimental case studies indicate, this semantic layer enables agents to catch geometric and volumetric errors in simulation prior to execution, recover statefully from mid-run hardware interruptions without wasting reagents, and autonomously design factorial diagnostics to isolate subtle physical biases such as fluid stratification and tip-wall carryover. By transforming wet-lab operations into auditable execution trajectories and structured trace dossiers, VERID can demonstrably separate strong systematic noise from small biological signals, which is the main physical prerequisite for autonomous agents to design, execute, debug, and learn from experiments at scale. The next step is demonstrating that these gains lead to significant biological breakthroughs on grand challenge research questions.

## 6. Methods

### 6.1. Robotic Workcell and Reagents

All experimental case studies were performed on an automated laboratory workcell centered on a Hamilton Microlab STAR liquid handler equipped with a 96-channel pipetting head and an 8-channel independent pipetting head (Span-8). Optical absorbance measurements were acquired on a Byonoy Absorbance 96 four-wavelength plate reader, with plate transport between the liquid handler and plate reader performed by a UFACTORY Lite 6 six-degree-of-freedom (6-DOF) robotic arm. Visual telemetry and overhead plate imaging were captured using consumer-grade RGB video cameras, including an OBSBOT Tiny 2 camera mounted on the liquid-handler gantry. All experiments used deionized water and aqueous solutions of retail food colorants (red, yellow, blue, and green) as model reagents.

### 6.2. Software Architecture and Agent Interface

VERID controls workcell hardware through Python device drivers built on the open-source PyLabRobot framework [25] and custom instrument interfaces. Autonomous agents interact with VERID remotely over a client–server architecture using a set of structured tool calls. To inspect the workcell, agents call workcell inspection/description tool, which returns a structured summary of the current semantic digital twin state, including labware positions, well volumes, and tip inventories. To test candidate protocols *in silico*, agents call a command verification tool, which executes a proposed command block against a forked virtual replica of the current digital twin state. Validated protocols are deployed to the physical hardware via command dispatch tool, which re-verifies the command block against the live digital twin state before initiating physical execution and returning a run identifier. Agents monitor execution and synchronize workflows using tools to poll for status and completion; upon run completion, synchronized trace dossier artifacts (state snapshots, execution logs, video feeds, and assay readouts) are written directly to the agent's workspace.

Agents are also provided with standard workspace file and code execution tools including search, analysis and mutation, along with a sandboxed shell execution tool restricted to the local workspace, tool outputs, and the persistent knowledge base. VERID is model-agnostic and decouples the workcell controller from the underlying reasoning model. Across all case studies, agents operated within a harness preloaded with structured instructions for protocol design and experiment iteration, a sandboxed Python environment with standard scientific computing libraries and static code analysis, and an operator approval gate locked to the verified script in supervised runs. Case Studies 1–3 were executed using an internal preview checkpoint of Gemini 4 Argon.

### 6.3. Scoring Rules for Case Study 3

Agents were scored on the composition value they output for each target mixture by scoring total dye concentration separately from relative dye concentration. For each target well, let the true and estimated (scored) constituent fractions in full-strength stock equivalents be the four dye fractions $f = (f_{\text{red}}, f_{\text{green}}, f_{\text{blue}}, f_{\text{yellow}}) \geq 0$ and the water fraction $w = f_{\text{water}} = 1 - \sum_{i=1}^{4} f_i$, with tempering exponent $\beta = 0.5$, smoothing constant $\varepsilon = 0.03$, water-fraction clip $w_{\text{clip}} = 10^{-4}$, and total-concentration weight $\lambda_{\text{total}} = \frac{\beta^2}{8} = 0.03125$.

**1. Relative Dye Shares ($s_i$):** With total dye fraction $F = \sum_{i=1}^{4} f_i$, the relative share of each dye is

$$s_i = \begin{cases} \frac{f_i}{F} & \text{if } F > 0 \ , \\ \frac{1}{4} & \text{if } F = 0 \ . \end{cases} \tag{1}$$

**2. Tempered Dye Distribution ($\tau_i$):** To assign sublinear weight to minor dyes and bound the penalty for omitting a present dye, relative shares are tempered with exponent $\beta$ and additive smoothing $\varepsilon$ before normalization:

$$\tau_i(f) = \frac{(s_i + \varepsilon)^\beta}{\sum_{j=1}^{4} (s_j + \varepsilon)^\beta}. \tag{2}$$

**3. Composition Loss ($L_{\text{comp}}$):** The relative composition error is the Kullback–Leibler divergence from the true tempered distribution $\tau^{\text{true}}$ to the scored distribution $\tau^{\text{scored}}$:

$$\begin{aligned} L_{\text{comp}} &= D_{\text{KL}}\big(\tau^{\text{true}} \parallel \tau^{\text{scored}}\big) \\ &= \sum_{i=1}^{4} \tau_i^{\text{true}} \ln\left(\frac{\tau_i^{\text{true}}}{\tau_i^{\text{scored}}}\right). \end{aligned} \tag{3}$$

**4. Total Concentration Loss ($L_{\text{total}}$):** The overall concentration error is the weighted squared logit error on the clipped water fraction $\tilde{w} = \text{clip}\big(w, w_{\text{clip}}, 1 - w_{\text{clip}}\big)$, where $\text{logit}(\tilde{w}) = \ln\big(\frac{\tilde{w}}{1-\tilde{w}}\big)$:

$$L_{\text{total}} = \lambda_{\text{total}}(\text{logit}\,\tilde{w}_{\text{scored}} - \text{logit}\,\tilde{w}_{\text{true}})^2. \tag{4}$$

**5. Target and Run Score:** Each target well's loss (in nats) is the sum of its composition and total concentration losses, and the overall run score is the mean across all four target wells (reported in millinats, $1\ \text{mnat} = 10^{-3}\ \text{nats}$):

$$\begin{aligned} \text{Score}_{\text{target}} &= L_{\text{comp}} + L_{\text{total}} \quad [\text{nats}], \\ \text{Score}_{\text{run}} &= \frac{1}{4} \sum_{k=1}^{4} \text{Score}_{\text{target},k}. \end{aligned} \tag{5}$$

## Acknowledgements

We thank all the members of Team Morphogenesis for their help and support throughout this study. We wish to thank Ronit Levavi Morad, Matthias Heiler and James Manyika for leadership support, and John Platt for reviewing the manuscript and providing valuable feedback. We thank Anthony Phalen for assistance on partnerships, Anna Sard for assistance on finance, Reya Shawky for building support and Amanie Brik for assistance on all the things to make a lab run fast.

## Author contributions

Y.D., P.S. and A.P. conceived and supported the study. Y.D. developed the lab controller, with J.C., Z.C., A.D., P.S., P.R. and A.P. providing critical support. M.B. and A.D. provided infrastructure support and feedback. J.W.K., O.D., J.C., and Z.C. provided user feedback, designed and executed assays for validation experiments. S.B.S. developed and integrated hardware drivers. T.C. and J.C. built a custom feedback system. S.K. provided critical feedback on the conceptual framing and reviewed the manuscript. A.H. and Y.M. provided research leadership and strategic guidance. P.S. and A.P. co-led the work. All authors contributed to writing, reviewing, and editing the manuscript.

## Competing interests

All authors performed this work while affiliated with Google LLC and were compensated for it by Google, and some of the authors may own Alphabet Inc. stock as part of their compensation package. The authors declare no other competing interests.

## References

[1] J. T. Rapp, B. J. Bremer, and P. A. Romero, "Self-driving laboratories to autonomously navigate the protein fitness landscape," *Nature Chemical Engineering*, vol. 1, no. 1, pp. 97–107, 2024, doi: 10.1038/s44286-023-00002-4.

[2] R. S. Bohacek, C. McMartin, and W. C. Guida, "The art and practice of structure-based drug design: a molecular modeling perspective," *Medicinal Research Reviews*, vol. 16, no. 1, pp. 3–50, 1996, doi: 10.1002/(SICI)1098-1128(199601)16:1<3::AID-MED1>3.0.CO;2-6.

[3] J.-L. Reymond, "The chemical space project," *Accounts of Chemical Research*, vol. 48, no. 3, pp. 722–730, 2015, doi: 10.1021/ar500432k.

[4] J. Maynard Smith, "Natural selection and the concept of a protein space," *Nature*, vol. 225, no. 5232, pp. 563–564, 1970, doi: 10.1038/225563a0.

[5] P. A. Romero and F. H. Arnold, "Exploring protein fitness landscapes by directed evolution," *Nature Reviews Molecular Cell Biology*, vol. 10, no. 12, pp. 866–876, 2009, doi: 10.1038/nrm2805.

[6] E. T. Boder and K. D. Wittrup, "Yeast surface display for screening combinatorial polypeptide libraries," *Nature Biotechnology*, vol. 15, no. 6, pp. 553–557, 1997, doi: 10.1038/nbt0697-553.

[7] B. J. Wittmann, Y. Yue, and F. H. Arnold, "Informed training set design enables efficient machine learning-assisted directed protein evolution," *Cell Systems*, vol. 12, no. 11, pp. 1026–1045, 2021, doi: 10.1016/j.cels.2021.07.008.

[8] J. Gottweis *et al.*, "Towards an AI co-scientist," *arXiv preprint arXiv:2502.18864*, 2025, doi: 10.48550/arXiv.2502.18864.

[9] J. Abramson *et al.*, "Accurate structure prediction of biomolecular interactions with AlphaFold 3," *Nature*, vol. 630, no. 8016, pp. 493–500, 2024, doi: 10.1038/s41586-024-07487-w.

[10] J. L. Watson *et al.*, "De novo design of protein structure and function with RFdiffusion," *Nature*, vol. 620, no. 7976, pp. 1089–1100, 2023, doi: 10.1038/s41586-023-06415-8.

[11] V. Zambaldi *et al.*, "De novo design of high-affinity protein binders with AlphaProteo," *arXiv preprint arXiv:2409.08022*, 2024, doi: 10.48550/arXiv.2409.08022.

[12] M. Pacesa *et al.*, "One-shot design of functional protein binders with BindCraft," *Nature*, vol. 646, no. 8084, pp. 483–492, 2025, doi: 10.1038/s41586-025-09429-6.

[13] F. Strieth-Kalthoff, F. Sandfort, M. Kühnemund, F. R. Schäfer, H. Kuchen, and F. Glorius, "Machine learning for chemical reactivity: the importance of failed experiments," *Angewandte Chemie International Edition*, vol. 61, no. 29, p. e202204647, 2022, doi: 10.1002/anie.202204647.

[14] J. Jumper *et al.*, "Highly accurate protein structure prediction with AlphaFold," *Nature*, vol. 596, no. 7873, pp. 583–589, 2021, doi: 10.1038/s41586-021-03819-2.

[15] L. Cao *et al.*, "Design of protein-binding proteins from the target structure alone," *Nature*, vol. 605, no. 7910, pp. 551–560, 2022, doi: 10.1038/s41586-022-04654-9.

[16] N. R. Bennett *et al.*, "Improving de novo protein binder design with deep learning," *Nature Communications*, vol. 14,

no. 1, p. 2625, 2023, doi: 10.1038/s41467-023-38328-5.

[17] M. Baker, "1,500 scientists lift the lid on reproducibility," *Nature*, vol. 533, no. 7604, pp. 452–454, 2016, doi: 10.1038/533452a.

[18] I. Holland and J. A. Davies, "Automation in the life science research laboratory," *Frontiers in Bioengineering and Biotechnology*, vol. 8, p. 571777, 2020, doi: 10.3389/fbioe.2020.571777.

[19] D. A. Boiko, R. MacKnight, B. Kline, and G. Gomes, "Autonomous chemical research with large language models," *Nature*, vol. 624, no. 7992, pp. 570–578, 2023, doi: 10.1038/s41586-023-06792-0.

[20] A. M. Bran, S. Cox, O. Schilter, C. Baldassari, A. D. White, and P. Schwaller, "Augmenting large language models with chemistry tools," *Nature Machine Intelligence*, vol. 6, no. 5, pp. 525–535, 2024, doi: 10.1038/s42256-024-00832-8.

[21] M. Abolhasani and E. Kumacheva, "The rise of self-driving labs in chemical and materials sciences," *Nature Synthesis*, vol. 2, no. 6, pp. 483–492, 2023, doi: 10.1038/s44160-022-00231-0.

[22] G. Tom *et al.*, "Self-driving laboratories for chemistry and materials science," *Chemical Reviews*, vol. 124, no. 16, pp. 9633–9732, 2024, doi: 10.1021/acs.chemrev.4c00055.

[23] N. Yoshikawa *et al.*, "Large language models for chemistry robotics," *Autonomous Robots*, vol. 47, no. 8, pp. 1057–1086, 2023, doi: 10.1007/s10514-023-10136-2.

[24] K. Darvish *et al.*, "ORGANA: A robotic assistant for automated chemistry experimentation and characterization," *Matter*, vol. 8, no. 2, p. 101897, 2025, doi: 10.1016/j.matt.2024.10.015.

[25] R. P. Wierenga, S. M. Golas, W. Ho, C. W. Coley, and K. M. Esvelt, "PyLabRobot: An open-source, hardware-agnostic interface for liquid-handling robots and accessories," *Device*, vol. 1, no. 4, p. 100111, 2023, doi: 10.1016/j.device.2023.100111.

[26] M. Grieves and J. Vickers, "Digital twin: Mitigating unpredictable, undesirable emergent behavior in complex systems," in *Transdisciplinary Perspectives on Complex Systems*, Springer, 2017, pp. 85–113. doi: 10.1007/978-3-319-38756-7_4.

[27] S. G. Baird and T. D. Sparks, "What is a minimal working example for a self-driving laboratory?," *Matter*, vol. 5, no. 12, pp. 4170–4178, 2022, doi: 10.1016/j.matt.2022.11.007.

[28] D. Ha and J. Schmidhuber, "Recurrent world models facilitate policy evolution," in *Advances in Neural Information Processing Systems (NeurIPS)*, 2018, pp. 2451–2463. doi: 10.5281/zenodo.1207631.

[29] D. Hafner, J. Pasukonis, J. Ba, and T. Lillicrap, "Mastering diverse control tasks through world models," *Nature*, vol. 640, no. 8059, pp. 647–653, 2025, doi: 10.1038/s41586-025-08744-2.

[30] Y. LeCun, "A path towards autonomous machine intelligence," *OpenReview preprint*, 2022, [Online]. Available: https://openreview.net/forum?id=BZ5a1r-kVsf

[31] D. Pathak, P. Agrawal, A. A. Efros, and T. Darrell, "Curiosity-driven exploration by self-supervised prediction," in *Proceedings of the 34th International Conference on Machine Learning (ICML)*, 2017, pp. 2778–2787. doi: 10.48550/arXiv.1705.05363.

[32] H. Tang, D. Key, and K. Ellis, "WorldCoder, a model-based LLM agent: Building world models by writing code and interacting with the environment," in *Advances in Neural Information Processing Systems (NeurIPS)*, 2024, pp. 70148–70212. doi: 10.48550/arXiv.2402.12275.

[33] P. A. Tsividis *et al.*, "Human-level learning of complex novel tasks as theory-based modelling, exploration and planning," *Philosophical Transactions of the Royal Society A: Mathematical, Physical and Engineering Sciences*, vol. 384, no. 2320, p. 20240529, 2026, doi: 10.1098/rsta.2024.0529.

[34] A. Zhao, D. Huang, Q. Xu, M. Lin, Y.-J. Liu, and G. Huang, "ExpeL: LLM agents are experiential learners," in *Proceedings of the AAAI Conference on Artificial Intelligence*, 2024, pp. 19632–19642. doi: 10.1609/aaai.v38i17.29936.

[35] Z. Zhang, H. Que, J. Chang, X. Zhang, H. Wei, and T. Zhu, "Safe-SDL: Establishing safety boundaries and control mechanisms for AI-driven self-driving laboratories," *arXiv preprint arXiv:2602.15061*, 2026, doi: 10.48550/arXiv.2602.15061.

[36] A. D. Ames, S. Coogan, M. Egerstedt, G. Notomista, K. Sreenath, and P. Tabuada, "Control barrier functions: Theory and applications," in *2019 18th European Control Conference (ECC)*, 2019, pp. 3420–3431. doi: 10.23919/ECC.2019.8796030.

[37] C. Baum *et al.*, "A system capable of verifiably and privately screening global DNA synthesis," *National Science Review*, vol. 13, no. 14, p. nwag103, 2026, doi: 10.1093/nsr/nwag103.